\documentclass[conference]{IEEEtran}

\usepackage{cite}
\usepackage{amsmath,amssymb,amsfonts}
\usepackage{algorithmic}
\usepackage{graphicx}
\usepackage{textcomp}
\usepackage{xcolor, soul}
\usepackage[hyphens]{url}
\usepackage{hyperref}
\def\BibTeX{{\rm B\kern-.05em{\sc i\kern-.025em b}\kern-.08em
    T\kern-.1667em\lower.7ex\hbox{E}\kern-.125emX}}

\usepackage{tikz}
\newcommand*\circled[1]{\tikz[baseline=(char.base)]{\node[shape=circle,fill,inner sep=0.5pt] (char) {\textcolor{white}{#1}};}}

\usepackage{mathtools}

\usepackage[ruled, linesnumbered, vlined]{algorithm2e}
\usepackage{setspace}
\SetAlFnt{\small}
\SetAlCapNameFnt{\small}
\SetAlCapFnt{\small}
\IncMargin{3pt}
\SetAlgoNlRelativeSize{-0.5}
\SetKwProg{Def}{def}{$\,$:}{}
\SetKwInOut{Input}{Input}
\SetKwInOut{Output}{Output}

\usepackage{etoolbox} 
\makeatletter
\patchcmd{\algocf@makecaption@ruled}{\hsize}{\textwidth}{}{} 
\patchcmd{\@algocf@start}{-1.5em}{0em}{}{} 
\makeatother

\usepackage{booktabs, makecell, multirow, tabularx} 
\usepackage{caption}
\begin{document}

\title{BBS: Bi-directional Bit-level Sparsity for \\ Deep Learning Acceleration
}

\author{
    \IEEEauthorblockN{Yuzong Chen}
    \IEEEauthorblockA{\textit{Cornell University}\\
    New York, NY, USA \\
    yc2367@cornell.edu}
\and
    \IEEEauthorblockN{Jian Meng}
    \IEEEauthorblockA{\textit{Cornell University}\\
    New York, NY, USA \\
    jm2787@cornell.edu}
\and
    \IEEEauthorblockN{Jae-sun Seo}
    \IEEEauthorblockA{\textit{Cornell University}\\
    New York, NY, USA \\
    js3528@cornell.edu}
\and
    \IEEEauthorblockN{Mohamed S. Abdelfattah}
    \IEEEauthorblockA{\textit{Cornell University}\\
    New York, NY, USA \\
    mohamed@cornell.edu}
}

\maketitle
\thispagestyle{plain}
\pagestyle{plain}

\begin{abstract}
Bit-level sparsity methods skip ineffectual zero-bit operations and are typically applicable within bit-serial deep learning accelerators.
This type of sparsity at the bit-level is especially interesting because it is both orthogonal and compatible with other deep neural network (DNN) efficiency methods such as quantization and pruning.
Furthermore, it comes at little or no accuracy degradation and can be performed completely post-training.
However, current bit-sparsity approaches lack practicality because of (1) load imbalance from the random distribution of zero bits, (2) unoptimized external memory access because all bits are fetched from off-chip memory, and (3) high hardware implementation overhead, including large multiplexers and shifters to support sparsity at the bit level.

In this work, we improve the practicality and efficiency of bit-level sparsity through a novel algorithmic bit-pruning, averaging, and compression method, and a co-designed efficient bit-serial hardware accelerator.
On the algorithmic side, we introduce bi-directional bit sparsity (BBS). 
The key insight of BBS is that we can leverage bit sparsity in a symmetrical way to prune either zero-bits or one-bits. 
This significantly improves the load balance of bit-serial computing and guarantees the level of sparsity to be more than 50\%. 
On top of BBS, we further propose two bit-level binary pruning methods that require no retraining, and can be seamlessly applied to quantized DNNs. 
Combining binary pruning with a new tensor encoding scheme, BBS can both skip computation and reduce the memory footprint associated with bi-directional sparse bit columns.
On the hardware side, we demonstrate the potential of BBS through \textit{BitVert}, a bit-serial architecture with an efficient PE design to accelerate DNNs with low overhead, exploiting our proposed binary pruning. 
Evaluation on seven representative DNN models shows that our approach achieves: (1) on average $1.66\times$ reduction in model size with negligible accuracy loss of $<0.5\%$; (2) up to $3.03\times$ speedup and $2.44\times$ energy saving compared to prior DNN accelerators.
\end{abstract}


\section{Introduction}
Deep neural networks (DNNs) have demonstrated remarkable accomplishments in many important fields such as computer vision and natural language processing. However, the growth of DNN model size and complexity continues to outpace the scaling of compute performance in existing hardware platforms~\cite{Gholami2024AIAM}. Bridging this performance gap is very desirable for wider adoption of DNNs, particularly in edge scenarios that demand both high performance and energy efficiency. 
Codesigning novel DNN compression algorithms, together with accelerators for the efficient deployment of the compressed models, is a promising way to achieve this goal. 

Numerous efficiency algorithms \cite{Molchanov2016PruningCN, mishra2021accelerating, lee2019snip} and hardware prototypes \cite{SparTen, dstc, gospa, eureka, highlight} have been proposed to leverage \textit{value-based sparsity} in DNNs to reduce the cost of storing and deploying DNNs. Yet the degree of such value sparsity, which depends on the underlying model architecture, can strongly limit the resulting hardware performance. For instance, 
recent transformer-based DNNs show limited or no activation sparsity with GeLU and sigmoid activation functions~\cite{vit, bert}. 
Even for single-sided sparse accelerators that target weight sparsity, plenty of time and cost are spent on retraining the model to balance the degree of sparsity and accuracy loss. 
Unfortunately, in many real-world cases, retraining may become impractical for end users due to cost constraints and lack of access to the original training dataset~\cite{raella, bitwave}. 
This challenge is particularly pronounced in recent large language models \cite{opt, llama} that contain billions of parameters, making retraining even more resource and data intensive.
Hence, there is a strong need to further enhance the efficiency of DNN accelerators \textit{without imposing retraining}.

Another line of DNN compression research focuses on \textit{post-training quantization} (PTQ), which represents DNN operands in lower precision without retraining the model \cite{vit_quant, noisyquant, ptq4vit, datafree_quant, smoothquant, olive, microscaling}. 
For example, researchers have designed new quantization data types such as the Microscaling format \cite{microscaling}, where a group of low-precision operands can share an 8-bit exponent to balance the accuracy and memory footprint. However, Microscaling still requires a floating-point pipeline to handle the shared 8-bit exponent, resulting in higher hardware cost than integer quantization. 
On the other hand, state-of-the-art PTQ algorithms can already reduce the operand precision to 8-bit integer with negligible accuracy loss~\cite{datafree_quant, noisyquant, smoothquant}. 
Unfortunately, a quantized 8-bit DNN shows extremely low value sparsity (less than 5\% as will be shown in the next section), since it tries to utilize all quantization levels as much as possible to reduce the quantization error. This fundamental quantization-sparsity tension poses a big performance bottleneck in existing value-based DNN accelerators~\cite{bitFusion, ant}. 

In order to jointly exploit the efficiency of quantization and sparsity, a series of bit-serial DNN accelerators exploit \textit{bit-level sparsity} \cite{stripes, pragmatic, bittactical, laconic, bitlet, bitwave}. Unlike coarse-grained value sparsity that is incompatible with quantization, the bit-level sparsity targets the abundant \textit{zero bits} in the binary representation of operands, thus is both compatible and orthogonal to other forms of DNN redundancy.
Stripes \cite{stripes} is an early bit-serial prototype that uses reduced precision for DNNs to scale the performance. Pragmatic \cite{pragmatic}, Laconic~\cite{laconic} and Bitlet \cite{bitlet} propose to skip zero-bit operations from different perspectives. 
However, the distribution of zero bits is generally random, whether in an individual operand or a group of operands, leading to significant workload imbalance. 
A direct consequence is that these accelerators must still fetch all data bits from off-chip memory, and use sophisticated hardware schedulers to skip zero-bit operations as much as possible during on-chip computation. The latter usually incurs non-trivial hardware overhead. 

To reduce both memory access and scheduling overhead of bit-serial computing, BitWave \cite{bitwave} employs a bit-column-serial approach, which examines the sparsity of the same bit significance across a group of operands. If a bit column contains all zero-bits, then it does not need to be stored in memory. Moreover, BitWave proposes a bit-sparsity-enhancing technique based on sign-magnitude formatted weights to selectively flip bits to zero. With this \textit{bit-flip} technique, BitWave is able to further compress a quantized 8-bit DNN by generating more zero bit columns. As a result, it has demonstrated the potential to achieve higher performance than other bit-serial accelerators \cite{stripes, pragmatic, bitlet}.

    \begin{figure}
        \centering
        \includegraphics[width=1\linewidth]{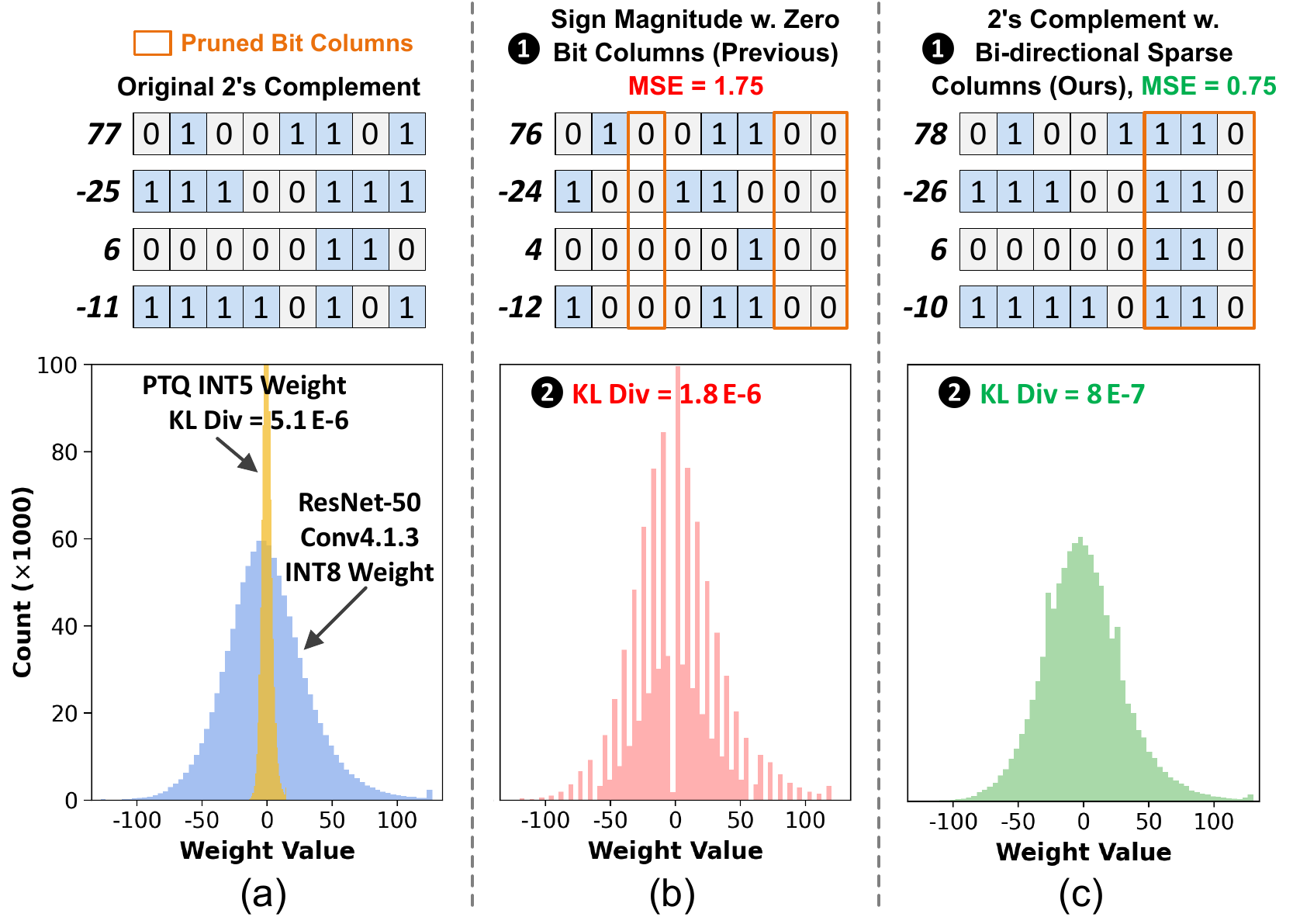}
        \vspace{-12pt}
        \caption{Comparison of different model compression approaches. (a) Example of a 4-value group and the weight distribution of a ResNet-50 layer before and after PTQ. (b) \protect\circled{\scriptsize1} Bit-sparsity enhancement by generating three zero bit columns using sign-magnitude format, \protect\circled{\scriptsize2}  achieving lower KL divergence than PTQ but still losing many quantization levels. (c) \protect\circled{\scriptsize1} BBS generates three bi-directional sparse bit columns and is able to preserve all quantization levels of 8-bit precision, \protect\circled{\scriptsize2} leading to much lower KL divergence.}
        \label{fig:weight_dist_comparison}
        \vspace{-8pt}
    \end{figure}

    \begin{figure*}
        \centering
        \includegraphics[width=1\linewidth]{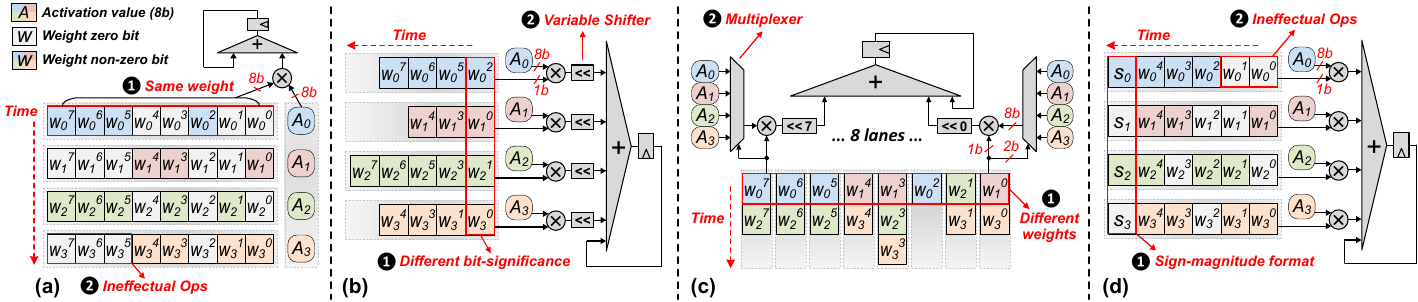}
        \vspace{-10pt}
        \caption{High-level computation flow of (a) bit-parallel PE, (b) Pragmatic \cite{pragmatic}, (c) Bitlet \cite{bitlet}, (d) BitWave \cite{bitwave}. }
        \label{fig:previous_works}
        \vspace{-8pt}
    \end{figure*}

Despite these approaches exploring bit sparsity at varying degrees, they still suffer from one significant drawback: bit sparsity is only limited to zero bits. 
To demonstrate this problem, consider Figure \ref{fig:weight_dist_comparison}(a) that shows a group of four INT8 values, as well as the INT8 weight distribution of a layer in ResNet-50. If we want to further reduce the bit-width to, \textit{e.g.}, 5-bit, conventional PTQ needs coarse-grained clipping and re-scaling so that the quantization mean square error (MSE) is minimized. Nevertheless, no matter what PTQ algorithm is used, the resulting distribution can only have $2^5 = 32$ discrete quantization levels, resulting in large KL divergence, a common metric to quantify the difference between two distributions \cite{kl_loss}. On the other hand, previous bit-sparsity-aware works \cite{dac_asbp, bittransformer, bitwave} leverage sign-magnitude format to prune bit columns at the group level as shown in Fig. \ref{fig:weight_dist_comparison}(b). Given that DNN weights are typically small, many inherent zero bit columns exist (\textit{e.g.}, the third bit columns in Fig. \ref{fig:weight_dist_comparison}(b)), leading to less sparse columns enforced (\textit{e.g.}, the seventh and eighth bit columns in Fig. \ref{fig:weight_dist_comparison}(b)) to achieve the effective 5-bit data width. As a result, they can preserve more quantization levels and achieve lower KL divergence and better accuracy than PTQ. However, if there is no inherent sparse bit column in a group, all lower significant bit columns must be flipped to zero, leading to reduced quantization levels especially in intervals with large absolute values (\textit{e.g.}, $>|50|$ in Fig. \ref{fig:weight_dist_comparison}(b)). 

\textbf{Our focus:} this work proposes a novel sparsity concept called \textit{bi-directional bit-level sparsity} (BBS) and the associate bit-serial accelerator design named \textit{BitVert}. The key insight of BBS is that the bit-level sparsity can be explored in a symmetrical way, where less zero-bits implies more one-bits, and vice versa. This ensures that any bit vector can exhibit at least $50\%$ BBS, which significantly improves the load balance of bit-serial computing while minimizing the number of ineffectual bit operations. Due to the balanced workload, BBS eliminates the expensive bit synchronization mechanism that is typically associated with prior bit-serial accelerators~\cite{pragmatic, bitlet, bittactical}. Furthermore, unlike previous bit-sparsity-aware works that only prune zero bit columns, BBS offers a new opportunity for model compression---it permits pruning a bit column with entirely zero-bits or entirely one-bits, which we call \textit{bi-directional sparse bit columns}. As shown in Fig. \ref{fig:weight_dist_comparison}(c), by looking for an optimal way to generate 3 bi-directional sparse columns, we can achieve much lower MSE compared to merely pruning zero bit columns with the same compression ratio. 
Additionally, since BBS allows any bit significance to be one, it preserves all quantization levels of the original INT8 weight and yields much lower KL divergence w.r.t. the original numerical distribution pre-compression. 
Finally, the balanced nature of BBS can be exploited in a hardware-friendly manner to improve the performance and energy efficiency of bit-serial accelerators. The main contributions of this work are summarized as follows:
\begin{enumerate}
    \item We introduce the new BBS concept, and demonstrate that
    BBS significantly improves the load balance of bit-serial accelerators. 
    \item We propose two bit-level \textit{binary pruning} strategies to enhance structured BBS. The binary pruning employs a new encoding scheme to reduce the memory footprint of a quantized DNN without the need of retraining.
    \item We design \textit{BitVert}, a bit-serial accelerator to exploit BBS for DNN acceleration. \textit{BitVert} adopts an efficient processing element (PE) with low hardware overhead for bit skipping, along with a channel-reordering mechanism to support binary pruning. 
\end{enumerate}

Through extensive evaluation on seven representative DNN benchmarks, including both vision and language models, we demonstrate that \textit{BitVert} achieves up to $3.03\times$ speedup and $2.44\times$ energy saving compared to prior DNN accelerators, while having negligible accuracy loss~($<\,$0.5\% on average) together with the preserved statistical characteristics of the uncompressed model.

\section{Background and Related Works}

\subsection{Sparse Bit-serial Accelerators}
We first describe the computation flow of bit-parallel processing and recent sparse bit-serial accelerators \cite{pragmatic, bitlet, bitwave} using a 4-way dot product example between 8-bit operands. We focus on weight sparsity in our discussion.
In Fig. \ref{fig:previous_works}(a), a bit-parallel PE exploits bit-level parallelism by performing the multiplication between an 8-bit activation and all bits of the same weight, but leading to many ineffectual bit operations. 
Since zero bits do not contribute to the final result, it is desirable to skip as many zero bits as possible for improved performance and efficiency. 

Pragmatic \cite{pragmatic} processes only non-zero bits of every weight as shown in Fig. \ref{fig:previous_works}(b). However, since different bit-significance can be processed simultaneously, Pragmatic requires a variable shifter after every bit-serial multiplier to synchronize the significance of essential bits. Bitlet \cite{bitlet} leverages the sparsity parallelism, motivated by the observation that every bit significance shows similar sparsity among a group of weights. As shown in Fig. \ref{fig:previous_works}(c), Bitlet digests multiple weights and activations, and computes every bit-significance independently. However, since every bit lane can absorb the essential bit from an arbitrary weight, Bitlet requires a large multiplexer (\textit{e.g.}, 64:1) to select the correct activation in every lane, leading to non-trivial hardware overhead ($35.9\%$ of the PE area as revealed by Bitlet's breakdown report). 
    \begin{figure}
        \centering
        \includegraphics[width=1\linewidth]{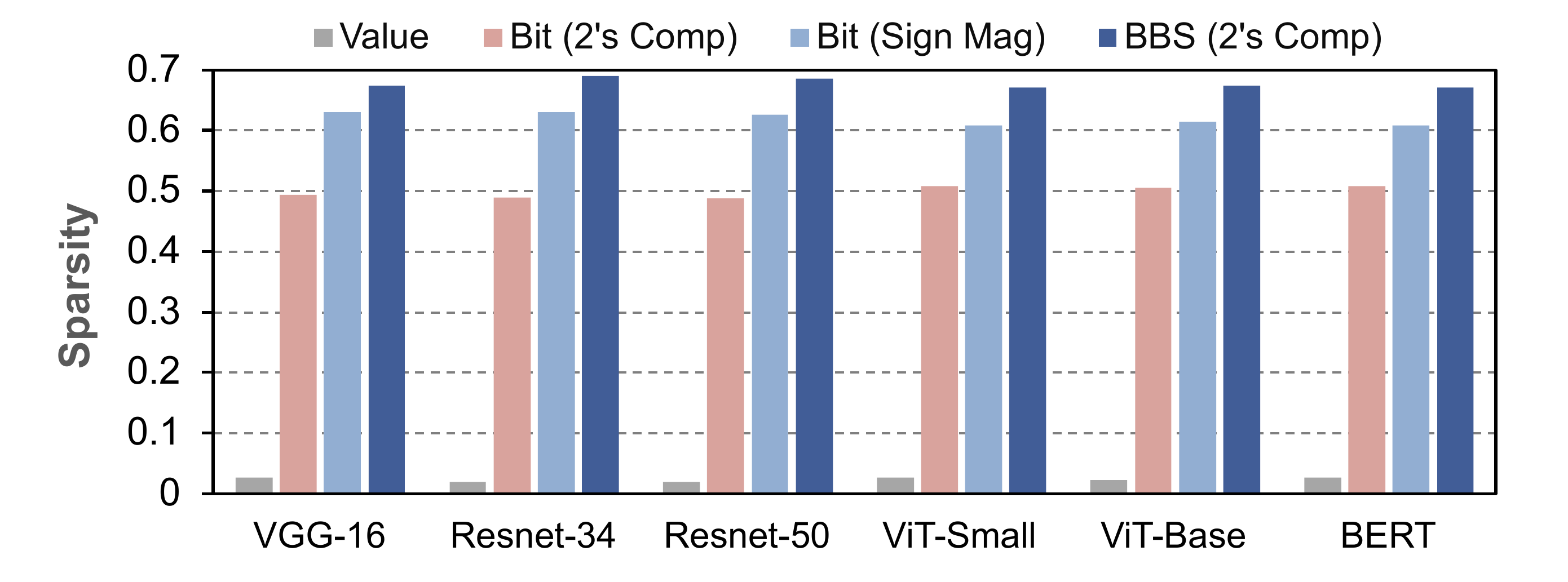}
        \vspace{-12pt}
        \caption{Comparison of inherent weight value sparsity, bit sparsity and BBS (with a bit-vector size of 8) in INT8 DNNs.}
        \label{fig:sparsity_comparison}
        \vspace{-8pt}
    \end{figure}

    \begin{figure*}
        \centering
        \includegraphics[width=1\linewidth]{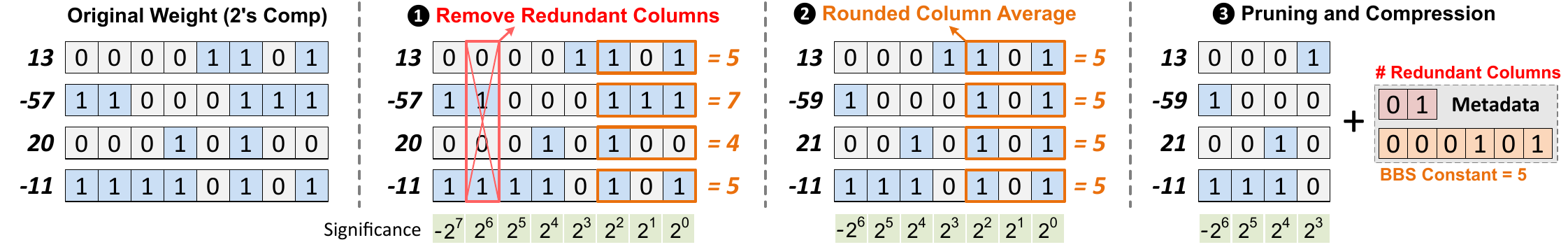}
        \vspace{-11pt}
        \caption{Example of bit-level binary pruning with rounded column averaging to generate 4 sparse bit columns.}
        \label{fig:bbs_round_avg}
        \vspace{-8pt}
    \end{figure*}
    
Both Pragmatic and Bitlet suffer from load imbalance issues, where the latency of Pragmatic is dominated by the weight with the highest number of one bits, and the latency of Bitlet is dominated by the bit significance with the highest number of one bits. To address this, BitWave \cite{bitwave} attempts to skip zero bits at the coarse bit-column granularity, as illustrated in Fig. \ref{fig:previous_works}(d). Because most weight values are typically small in a DNN, BitWave relies on sign-magnitude format which inherently generates many zero bit columns. The bit column sparsity offers balanced workload, but inevitably leads to many ineffectual bit operations since only a bit column with all zero bits can be skipped during computation. On the top of these three design philosophies, our proposed \textit{BitVert} tries to balance the bit-serial workload while skipping as many sparse bits as possible. 
By extending bit sparsity to BBS, \textit{BitVert} skips zero bits when a bit column contains many zeros, while it switches to skip one bits when a bit column contains less zero bits. Section \ref{sec:bbs} details our BBS methodology.

\subsection{Rethinking Bit-level Sparsity} \label{sec:rethink_bit_sparsity}
While recent advances in PTQ can compress DNNs to 8-bit with little or no accuracy loss \cite{zeroq, datafree_quant, smoothquant, vit_quant, ptq4vit}, the resulting weight tensor exhibits extremely low value sparsity. As shown in Fig. \ref{fig:sparsity_comparison}, the value-based weight sparsity is less than $5\%$ in a series of popular 8-bit quantized DNNs. This is because that a well-designed PTQ algorithm tries to utilize all available quantization levels to minimize the quantization MSE compared to original floating-point models. On the other hand, the bit-level sparsity is inherently more abundant and can achieve around $50\%$ in 2's complement format. Owing to the facts that DNN weight tensors usually exhibit Gaussian-like distribution and most values tend to be small \cite{olaccel, gobo, ant}, the sign-magnitude binary representation yields even higher bit sparsity \cite{bitwave, signmag_engine} due to abundant zero bits at higher bit significance. 
However, adopting sign-magnitude arithmetic for bit-serial computing still has two challenges. First, every bit-serial multiplier requires a 2's complementer for partial sum generation, resulting in large area overhead \cite{sibia}. 
Second, the irregular distribution of zero bits remains, leading to load imbalance and synchronization overhead. Whereas our proposed BBS maintains the 2's complement binary representation, and treats zero or one that has a higher occurrence as sparse bits. Hence, BBS ensures that any bit-vector exhibits at least $50\%$ bit sparsity, resulting in higher total bit sparsity than sign-magnitude format while achieving balanced workload across different PEs.


\section{BBS: Bi-directional Bit-level Sparsity} \label{sec:bbs}
    
In this section, we first introduce the concept of BBS based on 2's complement binary representation. Next, we present \textit{binary pruning}, a technique that modifies the original weight tensor to generate more structured BBS, together with a new encoding scheme that provides an extra opportunity for model compression. 
Finally, we propose a hardware-aware strategy to compress different weight channels of a DNN model based on the global awareness of pruning sensitivity, which can achieve favorable accuracy-compression trade-offs.  

\subsection{BBS Theorem} 
Without loss of generality, we describe BBS using a dot product operation that multiplies a group of $N$ weights ($W$) and activations ($A$) in $p$-bit precision, where $N$ is referred to as the \textit{group size}. In the rest of this paper, we use the term ``\textit{group}'' to refer to multiple weights or activations that contribute to the same dot product output. The dot product operation can be formally written as:
    \begin{equation}
        {\sum_{i=0}^{N-1} W_i \times A_i} \ = \ \\
        {\sum_{b=0}^{p-1} 2^b \times {\sum_{i=0}^{N-1} W_{i}^{b} \times A_i}}
        \label{eq:bit_serial_dp}
    \end{equation} 
where $W_{i}^{b}$ is the $b^{th}$ bit of $W_{i}$. Since any bit of $W$ can only be one or zero, the second partial sum on the right-hand side of Eq. \ref{eq:bit_serial_dp} can be re-organized as: 
    \begin{align}
        {\sum_{i=0}^{N-1} W_{i}^{b} \times A_i} \ &= \ \, {\sum_{\mathclap{\forall \, i : \, W_{i}^{b} = 1}}} \ {A_i} \label{eq:neg_bit_op} \\ 
        \ &= \ {\sum_{j=0}^{N-1} A_j} \, - \, {\sum_{\mathclap{\forall \, i : \, W_{i}^{b} = 0}}} \  {A_i}
        \label{eq:pos_bit_op}
    \end{align}

From Eq. \ref{eq:neg_bit_op} and \ref{eq:pos_bit_op}, we can infer that instead of adding the effectual activations associated with non-zero weight bits, the same result can be obtained by subtracting the activations indicated by zero weight bits from the sum of all activations, which is a constant for a given group.
Since more zero-bits in a vector implies less one-bits, Eq. \ref{eq:neg_bit_op} and Eq. \ref{eq:pos_bit_op} always process no more than half of the bits---when there are more than $50\%$ zero-bits in a bit-vector, the computation can skip them as in conventional bit-serial accelerators. But if there is less than $50\%$ bit sparsity, the bit-vector can be \textit{inverted} so that the original one-bits become sparse, and subtract the bit-serial dot product from $\sum_{i=0}^{N-1}A_i$. Since both zero and one can become sparse bits, we call this \textit{bi-directional bit sparsity (BBS)}.

The idea of BBS can effectively improve the load balance of bit-serial computing. Although there is ${\sim}50\%$ zero bit sparsity in 2's complement format and more than $50\%$ zero bit sparsity in sign-magnitude format (Fig. \ref{fig:sparsity_comparison}), the sparsity within a bit-vector is unpredictable. Moreover, because bit-serial computing relies on strongly increased parallelism to simultaneously process many bit-vectors from different weight groups, any bit-vector with low zero bit sparsity will hamper the performance of the whole PE array. On the other hand, BBS ensures at least $50\%$ sparsity in a bit-vector of arbitrary length, achieving balanced workload during parallel execution while skipping as many ineffectual bit operations as possible. 

\subsection{Bit-level Binary Pruning} \label{sec:binary_pruning}
In addition to balanced bit sparsity, BBS offers a new opportunity for model compression through \textit{binary pruning}---which can prune a bit column that contains all zero-bits or all one-bits within a weight group. Specifically, Eq. \ref{eq:neg_bit_op} implies that if all weight bits at a bit significance are zero, then the bit-serial dot product at that significance is simply zero. Similarly, Eq. \ref{eq:pos_bit_op} implies that if all weight bits at a significance are one, then the bit-serial dot product at that significance is the sum of activations in the group. As a result, a bi-directional sparse bit column can be compressed to just one bit that indicates whether its bit-serial dot product produces zero or sum of activations. Based on this observation, we propose two BBS-enhancing strategies to generate more bi-directional sparse bit columns in the original weight group, which can be effectively pruned through a new encoding scheme. 

    \begin{figure*}
        \centering
        \includegraphics[width=1\linewidth]{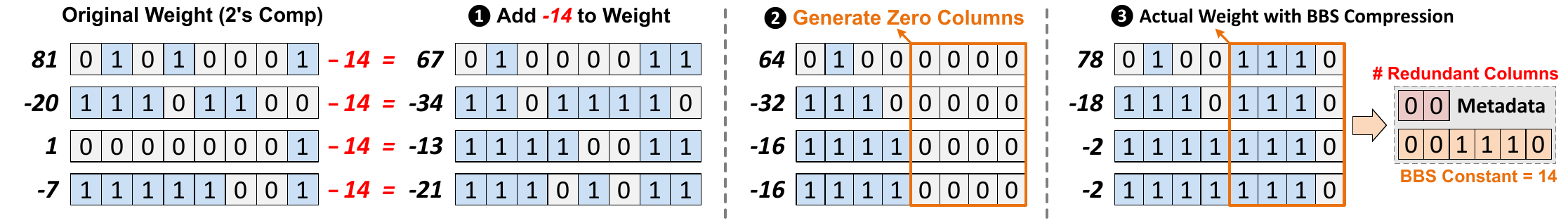}
        \vspace{-12pt}
        \caption{An example of bit-level binary pruning with zero-point shifting to generate 4 sparse bit columns.}
        \label{fig:bbs_zp_shifting}
        \vspace{-8pt}
    \end{figure*}

\vspace{5pt}
\noindent
\textbf{BBS with Rounded Averaging }
Fig. \ref{fig:bbs_round_avg} describes the procedure of the first BBS-enhancing strategy, \textit{rounded averaging}, using a group of 4 weights. Given the target number of sparse bit columns (4 in this example), Step \protect\circled{\small1} identifies if there are \textit{redundant} bit columns that immediately follow the most-significant column with the same content (e.g., the second bit column). 
Removing the redundant columns does not affect the original weight values as long as the remaining bits are interpreted as 2's complement format. For instance, the decimal number $-57$ in 8-bit 2's complement format is $11000111_b$, where the most-significant bit is multiplied by $-2^7$. Removing the second bit leads to a 7-bit number $1000111_b$, which is still equal to $-57$ if the most-significant bit is multiplied by $-2^6$.
After pruning the redundant column, the required number of bi-directional sparse columns to be generated is 3. These sparse columns are always generated from the lower significant bits, since modifying higher bit significance will increase the MSE exponentially. In Step \protect\circled{\small2}, this is achieved by calculating the rounded average of the values represented by the 3 lower significant bits of original weights. Essentially, this is replacing the 3 lower significant bits of all weights with a 3-bit constant while minimizing the MSE. Finally, Step \protect\circled{\small3} compresses the original weight group by storing only the remaining 4 bit columns and an 8-bit encoding metadata. 

\vspace{5pt}
\noindent
\textbf{BBS Compression Encoding } 
The encoding metadata contains 2 bits to specify the number of redundant columns, which can vary from 0 to 3, and 6 bits to store the BBS constant. The size of the metadata is chosen empirically. First, although there may be more than 3 redundant columns in a group, we find that this probability is extremely low for a large group size (\textit{e.g.}, 32) which amortizes the cost of metadata. If there are more than 3 redundant columns, we simply prune the first 3 and average additional lower significant columns instead. Second, using more than 6 bits to store the constant is also unnecessary since pruning 7 columns of an 8-bit tensor leaves only one effective bit, while pruning 8 columns means replacing all weights with the same 8-bit constant. Both situations can lead to unacceptable accuracy loss. 

    
\vspace{5pt}
\noindent
\textbf{BBS with Zero-point Shifting } 
The rounded column averaging strategy is particularly suitable for pruning a small number of bit columns, where the lower significant bits within a group are likely to have similar values. However, for more eager compression, \textit{i.e.}, pruning many columns, simply taking the rounded average over many lower significant bits of a group may lead to large MSE. Here is a simple example: assume we want to average only the least significant bit within a group of weights, then some weights will have no error after rounded averaging. On the other hand, if we average 4 lower significant bits, then all weights may produce error since any weight can have a different value in the 4 lower significant bits. 

To address this, we propose a second BBS-enhancing strategy called \textit{zero-point shifting}. The idea is to add an optimal constant to the original weight group (\textit{i.e.}, shifting its zero-point), which in turn facilitates the generation of sparse bit columns in the new weight group while minimizing the MSE. Fig. \ref{fig:bbs_zp_shifting} exemplifies this procedure for generating 4 sparse bit columns. In Step \protect\circled{\small1}, assume a constant $-14$ is added to the original weight, which changes the binary content of all numbers. Fortunately, the change of binary content makes it easier to generate zero columns in lower significant bits. 
As shown in Step \protect\circled{\small2}, to minimize the MSE when pruning the 4 lower significant bit columns, a number can either directly zero out the 4 lower bits (e.g., the first number changes from $67$ to $64$), or round up to the higher bit significance (e.g., the second number changes from $-34$ to $-32$).
Finally, Step \protect\circled{\small3} shows the actual values after binary pruning and stores the new zero-point in the encoding metadata.

\begin{algorithm} [t]
    \caption{Finding the optimal constant for zero-point shifting.}
    \label{algo:zp_shifting}
    \setstretch{1.05}
    
    \Input{Weight group: $W$, BBS constant precision: $p$, \\
    target number of sparse bit columns: $N$}
    \Output{Compressed weight: $W_C$, metadata: $D$}
    
    \Def{\upshape Compress($W$, $N$, $p$)}{
        bestMSE = $\infty$ \;
        \For{\upshape constant = $-2^{p-1}$ \textbf{to} $2^{p-1} - 1$} {
            $W_{tmp}$ = Clip$\,(W\, +$ constant$\,)$ \\
            \tcp{Get number of redundant columns}
            numRedunCol = GetNumRedunCol$(W_{tmp})$ \\ 
            $W_{tmp}$ = RemoveRedunCol$(W_{tmp}\,$, numRedunCol$\,)$ \\
            
            \tcp{Generate zero sparse columns}
            numSparseCol = $N\, -$ numRedunCol \\
            $W_{tmp}$ = GenSparseCol$(W_{tmp}\,$, numSparseCol$\,)$ \\
            newMSE = $| W_{tmp} - W |^{2} $ \\
            \If{\upshape newMSE $ < $ bestMSE} {
                bestMSE = newMSE \\
                $W_C$ = $W_{tmp}$ \\
                $D$ = \{$\,$numRedunCol$\,$, constant$\,$\} \\
            } 
        }
        \Return{$W_C$, $D$} \\ 
    }
\end{algorithm}

Algo. \ref{algo:zp_shifting} details the algorithm to find the optimal BBS constant for a weight group. 
Given the precision of the constant (6-bit in our proposed BBS encoding), the algorithm iterates through all possible constants (Line 3). In every iteration, it adds the current constant to the original weight group, followed by clipping to avoid overflow (Line 4). Next, similar to \textit{rounded averaging}, we calculate the number of redundant columns, and generate required number of sparse columns while minimizing MSE (Line 5 -- 7). Since the best constant will be stored in the BBS constant region of the metadata, we only generate zero sparse bit columns (Line 8) so that no extra encoding information is needed. Lastly, the algorithm checks whether the current constant results in lower MSE and updates the weight group and metadata accordingly (Line 9 -- 13). 

Although Algo. \ref{algo:zp_shifting} describes the procedure using a single weight group, the whole algorithm can be vectorized to find the optimal constant of all groups within a DNN layer simultaneously. During real implementation, the algorithm takes several milliseconds to several seconds per layer (totally ${\sim}$15s to compress the whole ResNet50) on a single Nvidia RTX 3090 GPU.
Hence, the proposed bit-level binary pruning method exhibits high efficiency and fast compression compared to prior quantization-oriented algorithms \cite{zeroq, ptq4vit, gobo}.

\begin{figure}
        \centering
        \includegraphics[width=1\linewidth]{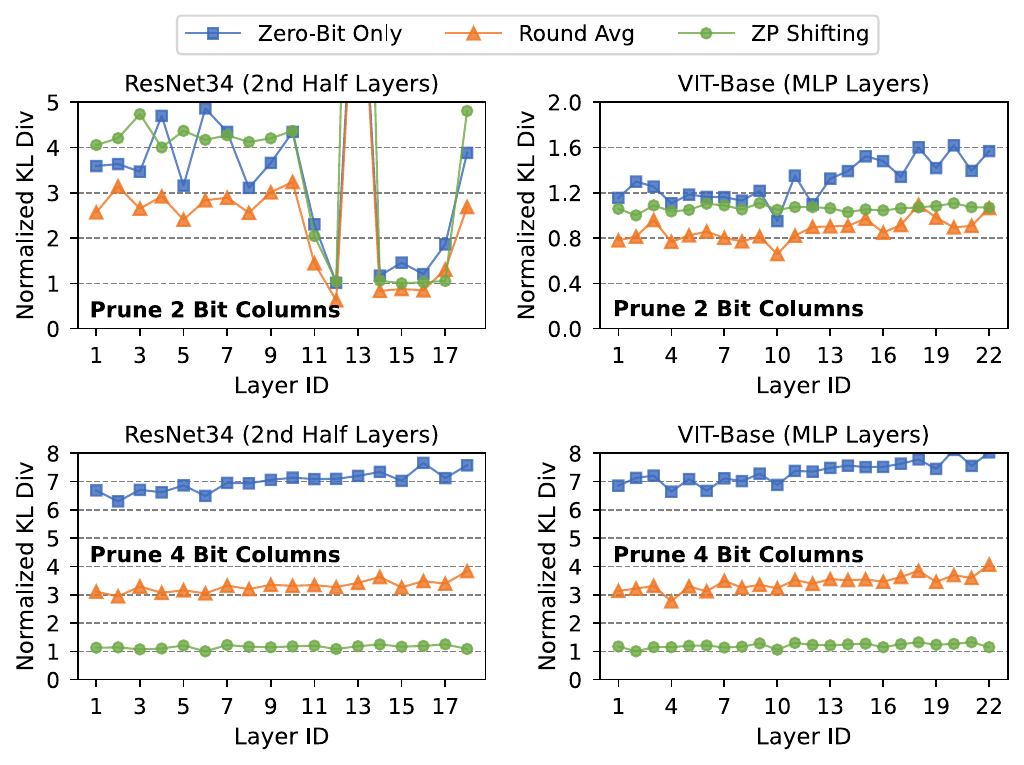}
        \vspace{-15pt}
        \caption{Normalized KL divergence (lower is better) of different bit-level pruning techniques with a weight group size of 32. }
        \label{fig:kl_div_comparison}
        \vspace{-5pt}
    \end{figure}

\vspace{5pt}
\noindent
\textbf{Rationality of Binary Pruning } \label{sec:e}
To demonstrate the rationality of the proposed two binary pruning strategies compared to previous zero-bit-only pruning \cite{bitwave, bittransformer, dac_asbp}, we apply the three techniques to compress the quantized 8-bit ResNet-34 and ViT-Base. Fig. \ref{fig:kl_div_comparison} shows the resulting KL divergence of different methods after pruning 2 and 4 bit columns with a weight group size of 32. The KL divergence is a common metric to quantify the difference between two distributions \cite{kl_loss, quant_survey}. A lower KL divergence indicates that the compressed weight tensor can better preserve the information of the original 8-bit weight, thus achieving better inference accuracy (evaluated in Section \ref{sec:eval_accuracy}). 

Specifically, Fig. \ref{fig:kl_div_comparison} shows that when pruning 2 bit columns, \textit{rounded averaging} consistently outperforms other approaches. The reason is that different weights within a group are likely to have similar values in the lower significant bits. On the other hand, \textit{zero-point shifting} yields much lower KL divergence when pruning 4 bit columns. This is because it can better exploit the binary characteristics of a weight group to find the optimal zero point that facilitates the generation of more sparse bit columns. Furthermore, the proposed binary pruning permits the existence of both zero and one in any bit significance after compression, thus are able to preserve all quantization levels of the original 8-bit weights as opposed to zero-bit-only pruning. As a result, both of our strategies show significant improvements when applied to a large number of bit columns. 

\subsection{Hardware-aware Global Binary Pruning} \label{sec:global_pruning}
So far, we have described binary pruning at the group level.
In order to fully exploit the structured BBS sparsity induced by binary pruning while mitigating the accuracy loss for the whole DNN, we propose a hardware-aware global binary pruning approach at the \textit{per-channel} granularity. 
Specifically, we find that the pruning sensitivity of different weight channels can be effectively quantified through magnitude-based proxies. 
For example, in convolutional neural networks, the sensitive filters (\textit{i.e.,} weight channels) usually contain many outliers with large magnitude. More specifically, in per-channel quantized DNNs, the sensitive channels of a weight tensor will have large scaling factors to accommodate these outliers \cite{smoothquant, perchannel_quant}. 
The per-channel weight quantization has been widely adopted to achieve high accuracy in state-of-the-art DNN accelerators \cite{raella, ant} and acceleration frameworks such as TensorRT \cite{tensorRT}. Therefore, we consider per-channel quantized 8-bit DNNs as the baseline for global binary pruning 
\footnote{For 8-bit DNNs that do not use per-channel quantization, other channel importance proxies such as the standard deviation of a weight channel can also be used to identify sensitive channels. }. 

\begin{algorithm} [t]
    \caption{Global binary pruning.}
    \label{algo:global_pruning}
    \setstretch{1.05}
    
    \Input{Model: $M$, per-channel scaling factors: $S$ \\
        threshold: $\beta$, hardware parameter: $C_H$}
    \Output{Pruned model: $M_P$}
    
    \Def{\upshape GlobalPrune($M$, $S$, $\beta$, $C_H$)}{
        \tcp{Global channel sorting}
        channelSorted = SortChannel$(M.channel, \,S\,)$ \\
        sensChannel = channelSorted$\,[\,1\,$ : $\,\beta \times Length(S)\,]$ \\
        \For{\upshape $L$ in $M.layers$} {
            \tcp{Ensure every layer has a multiple of $C_H$ sensitive channels}
            layerChannel = SortChannel$(L.channel, \,S[L]\,)$ \\
            numSens = Count$(\,$layerChannel $ \cap $ sensChannel$\,)$ \\
            numSens = Ceiling$(\,$numSens / $C_H\,) \times C_H$ \\
            \tcp{Get sensitive channels of layer $L$}
            
            topChannel = layerChannel$\,[\,1\,$: numSens$\,]$ \\
            sensChannel = sensChannel $\cup$ topChannel \\
        } 
        normalChannel = $M.channel \, - $ sensChannel \\
        \uIf{\upshape eagerCompression} {
            $M_P$ = RoundedAveraging$($normalChannel$)$ 
        } \Else {
            $M_P$ = ZeroPointShifting$($normalChannel$)$
        }
    \Return{\upshape $M_P$} 
    }
\end{algorithm}

\begin{figure*}
    \centering
    \includegraphics[width=1\linewidth]{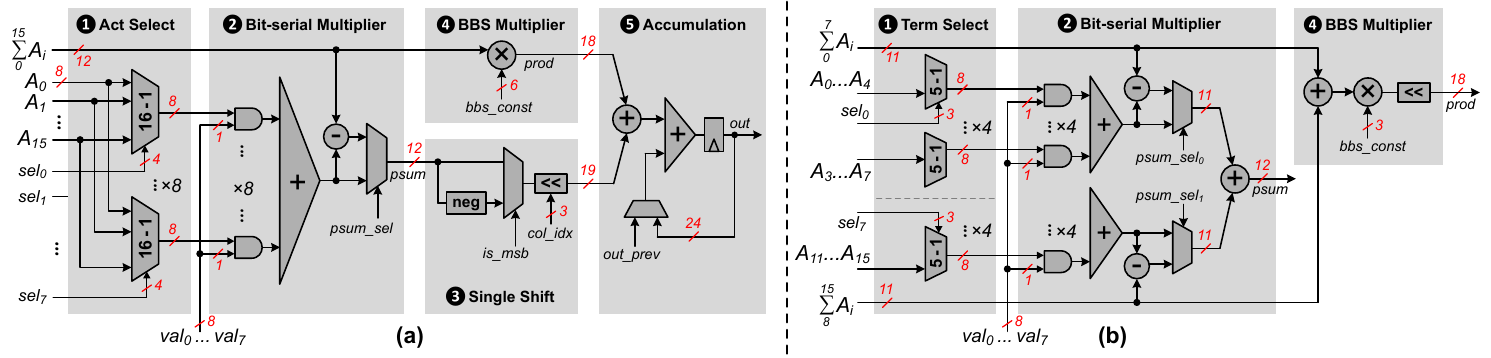}
    \vspace{-12pt}
    \caption{BitVert PE: (a) baseline design, (b) modified design.}
    \label{fig:bitvert_pe}
    \vspace{-8pt}
\end{figure*}

To apply global binary pruning, we define a hyperparameter $\beta$ to specify the minimum percentage of sensitive weight channels. Also, we define a hardware-aware parameter $C_H$, which specifies the number of weight channels processed in parallel during hardware acceleration (\textit{e.g.}, $C_H = 32$ in our \textit{BitVert} accelerator). Algo. \ref{algo:global_pruning} details the procedure of global binary pruning. The algorithm starts with global channel sorting to identify $\beta$ sensitive channels based on the scaling factors (Line 1 -- 2). For every layer, we force the number of sensitive channels to be a multiple of $C_H$ (Line 4 -- 9). For example, in the convolution layer, if the number of sensitive filters is less than $C_H$ after global channel sorting, then we simply select $C_H$ filters with the highest scaling factors as new sensitive channels. Finally, we apply binary pruning to the remaining channels (Line 10 -- 14), which can either prune a different number of bit columns for different layers~\cite{bitwave} or prune the same number of bit columns for all layers. 

The identification of sensitive channels further reduces the MSE and KL divergence while eliminating the need for resource-intensive and time-consuming retraining. 
In most of our DNN benchmarks (Section \ref{sec:methodology}), we are able to set $\beta = 10\%$ or $20\%$ while pruning a large number of bit columns in the remaining channels. However, since the locations of sensitive channels are random within a layer, two challenges arise for efficient hardware acceleration. First, identifying the location of sensitive channels requires significant indexing overhead. Second, different precision will cause unaligned memory access to the weight tensor in DRAM. The proposed \textit{BitVert} accelerator addresses these challenges through a channel-reordering mechanism as will be discussed shortly.



\section{BitVert Hardware Architecture} \label{sec:bitvert}
To fully exploit the potential of BBS and binary pruning, we design a bit-serial accelerator, named \textit{BitVert}, which
includes an efficient PE and scheduler to support BBS with compression, along with the channel reordering mechanism for hardware-aware global binary pruning.

\subsection{BitVert Processing Element}  \label{sec:bitvert_pe}
The \textit{BitVert} PE performs bit-serial multiplication between a group of 16 weights and activations, where weights are processed bit-serially. Fig. \ref{fig:bitvert_pe}(a) shows a baseline \textit{BitVert} PE that performs the computation in 5 steps. 
Step \protect\circled{\small1} receives 16 activations $A_0, ..., A_{15}$ and selects 8 of them based on $sel_0, ..., sel_{7}$ that indicates the position of effectual bits in the weight bit-vector. 
Step \protect\circled{\small2} performs bit-serial multiplication using valid signals $val_0, ... , val_7$ in case there are less than 8 effectual bits (\textit{i.e.}, more than $50\%$ sparsity in the weight bit-column). A subtractor subtracts the adder tree result from the sum of activations (Eq. \ref{eq:neg_bit_op}), followed by a mux to select the partial sum. Step \protect\circled{\small3} then shifts the partial sum based on the column index $col\_idx$ that specifies the significance of current weight bits. The $col\_idx$ can vary across different groups according to the number of redundant columns during binary pruning (Section \ref{sec:binary_pruning}). Recall that BBS compression stores a constant, whose ``0'' bit indicates a bit-column of all zero-bits and ``1'' bit indicates a bit-column of all one-bits. Hence, Step \protect\circled{\small4} multiples this constant with the sum of activations. Finally, the product and bit-serial partial sum are accumulated in Step \protect\circled{\small5}. The activations are reused for multiple clock cycles until all bit-columns belonging to the same weight group are processed. 
The control signals such as $sel$, $val$, and $col\_idx$ are updated by the \textit{BitVert} scheduler in every cycle (described in Section \ref{sec:bitvert_scheduler}).

Due to the random distribution of effectual bits within a weight bit-column, the baseline PE accounts for the worst case by using a 16:1 mux for every activation term. Since BBS guarantees at least $50\%$ sparsity in a bit-vector of arbitrary length, it is possible to reduce the mux cost with a smaller group size. 
Based on this observation, we propose a modified PE that computes bit-serial multiplication within a smaller \textit{sub-group} as shown in Fig. \ref{fig:bitvert_pe}(b). The sub-group size is a design parameter that offers a trade-off between area and power. A smaller sub-group can reduce the mux cost but requires more subtractors. Therefore, we conduct a PE design space exploration (Section \ref{sec:hardware_explore}) and choose a sub-group size of 8 in our design. Furthermore, because the PE supports $50\%$ bit sparsity, at most 4 activations will be selected within a sub-group. In the worst case, the selected activations within the sub-group $\{A_0, ..., A_7\}$ will be $\{A_4, A_5, A_6, A_7\}$. Hence, we only need four 5:1 muxes to locate all effectual activations, where the first mux selects among $\{A_0, ..., A_4\}$, the second mux selects among $\{A_1, ..., A_5\}$, and so on. Using 5:1 muxes further reduces the PE area compared to 8:1 muxes. 

It is also possible to reduce the cost of the BBS multiplier in Step \protect\circled{\small4}. Since BBS can prune a maximum of 6 bit columns in a weight group (Section \ref{sec:binary_pruning}), it requires at least 2 cycles to process the remaining columns when the weight precision is 8 bits. 
This allows time-multiplexing the BBS multiplier by multiplying 3 bits per cycle, followed by a shifter to align the significance. Section \ref{sec:hardware_explore} evaluates the reduction in PE area overhead achieved by the proposed optimization.

    \begin{figure}
        \centering
        \includegraphics[width=1\linewidth]{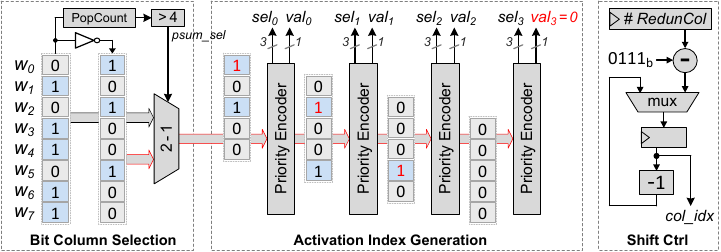}
        \vspace{-12pt}
        \caption{BitVert scheduler.}
        \label{fig:bitvert_scheduler}
        \vspace{-8pt}
    \end{figure}

\subsection{BitVert Scheduler} \label{sec:bitvert_scheduler}
\textit{BitVert} adopts a low-cost scheduler to control the operation within a PE, as illustrated in \text{Fig. \ref{fig:bitvert_scheduler}}. 
To control the bit-serial dot product, the scheduler first identifies whether there are more zero bits in a bit column. 
It then sends the original or inverted bit column to a series of 4 priority encoders. Every priority encoder receives 5 consecutive bits from the weight bit column. For example, the first priority encoder receives $\{w_0,..,w_4\}$, the second receives $\{w_1,..,w_5\}$, and so on. The encoder detects the location of the first ``1'' bit in the received 5-bit vector. If exists, it will mask the detected ``1'' bit and sends the remaining bits to the next encoder. On the other hand, if the received 5-bit vector contains all zero-bits, the encoder will signal $val = 0$ to disable the corresponding bit-serial multiplier in the PE. 

The scheduler also generates the $col\_idx$ signal to control the shifting of bit-serial multiplier in every PE. When a new dot product begins, the scheduler receives the BBS metadata which contains the number of redundant columns, $\#RedunCol$, in a weight group. The highest bit significance of the compressed weight group indicates the initial $col\_idx$ and is obtained by subtracting the number of redundant columns from 7 (i.e., the highest bit significance of uncompressed weight). The $col\_idx$ is updated in every cycle by subtracting one until the bit-serial bot product completes.

\subsection{Channel Reordering} \label{sec:channel_reorder}
\begin{figure} 
        \centering
        \includegraphics[width=1\linewidth]{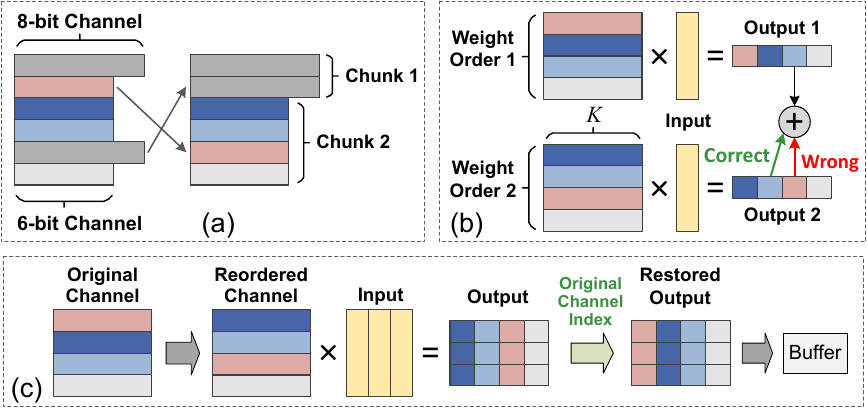}
        \vspace{-12pt}
        \caption{Channel reordering: (a) Store channels with the same precision in the same memory chunk. (b) Two weight tensors in a residual block with different channel orders can lead to the wrong result when processing the same input. (c) Unshuffle the output to restore the original channel order.}
        \label{fig:channel_reorder}
        \vspace{-10pt}
    \end{figure}

With per-channel global binary pruning, the sensitive and normal channels will have different precision, resulting in unaligned memory layout. To address this issue, we adopt a channel reordering mechanism as shown in Fig. \ref{fig:channel_reorder}(a). There are 6 weight channels in this example, and channels with the same precision are grouped together and stored in a memory chunk to avoid unaligned access. Recall from Section \ref{sec:global_pruning} that the proposed global binary pruning is hardware-aware, which forces the number of sensitive channels in every layer to be a multiple of the number of channels processed in parallel. Therefore, the grouped channels can be efficiently accessed by \textit{BitVert} to ensure full hardware utilization.  

The channel reordering mechanism has also been explored in SparTen's greedy balancing \cite{SparTen}. However, the reordering criteria is completely different. SparTen is a value-based sparse DNN accelerator that reorders weight channels based on their sparsity, while \textit{BitVert} groups channels based on their sensitivity to binary pruning. Furthermore, SparTen statically unshuffles the next layer’s weights in software, which may not guarantee the correctness when different weight tensors need to process the same input. Consider the example shown in \text{Fig. \ref{fig:channel_reorder}(b)}, where two weight tensors multiply the same input and generate two output tensors that require element-wise addition (\textit{e.g.}, as in the residual block of ResNet). SparTen statically unshuffles the two weight tensors along the $K$-dimension to align with the channel order of the previous layer, but the different channel order between the two weight tensors remains, which produce two output tensors with different orders. In this example, the second element of output 2 is supposed to be added with the third element of output 1, while a conventional design like SparTen will add the same position of two output tensors, leading to the wrong result. 

To solve the above issue, we propose to unshuffle the output tensor when writing back to memory. As shown in Fig. \ref{fig:channel_reorder}(c), after completing the whole dot product between the input tensor and reordered weight, the outputs are directly restored to the original channel order. This restoring only needs to know the original index of every weight channel to calculate the corresponding memory address for storing the final outputs. Fortunately, since a weight channel usually contains hundreds to thousands of values, the overhead of storing one index per channel is trivial. Moreover, because the same weight channel can process many inputs (3 in this example) to compute many outputs simultaneously, these outputs can be unshuffled together to amortize the cost of channel reordering.

    \begin{figure}
        \centering
        \includegraphics[width=1\linewidth]{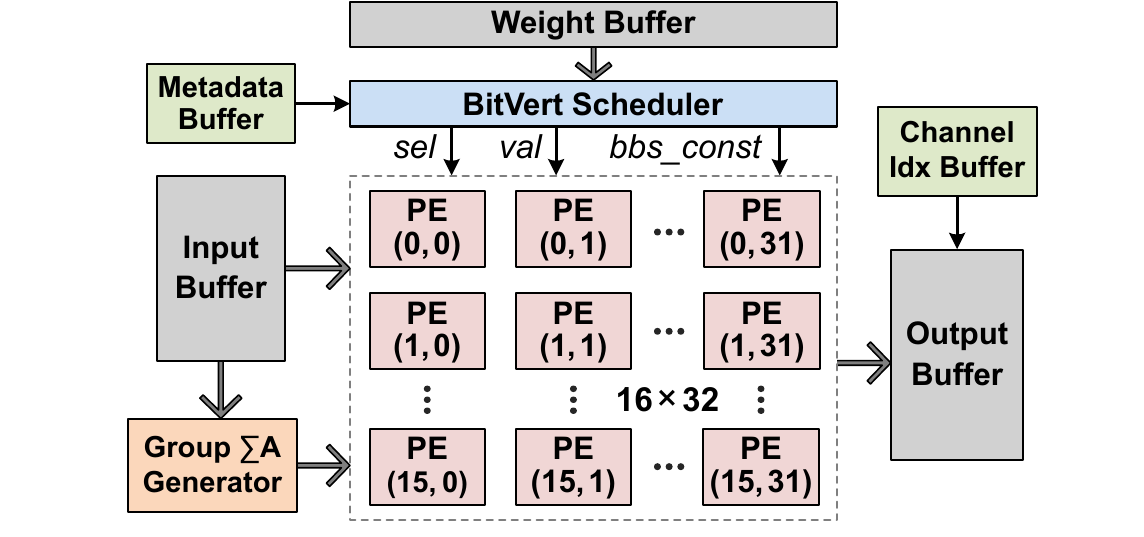}
        \vspace{-12pt}
        \caption{BitVert accelerator.}
        \label{fig:bitvert_accelerator}
        \vspace{-10pt}
    \end{figure}
    
\subsection{BitVert Accelerator} 
Fig. \ref{fig:bitvert_accelerator} shows the overall architecture of the \textit{BitVert} accelerator. The $16 \times 32$ PE array adopts an output-stationary dataflow, and exploits both weight-sharing and input-sharing by processing 32 weight channels and 16 input windows in parallel. The weight and input buffers are banked to provide adequate bandwidth for the access from PEs. Outputs are read out of the PE array and written to the output buffer, one column at a time. Additionally, \textit{BitVert} incorporates a metadata buffer to store BBS compression metadata, and a channel index buffer to store the original index of weight channels being processed. The $\Sigma A$ generator calculates the sum of input activations for BBS-based bit-serial multiplication inside the PE. Since the same input group is multiplied by 32 weight channels, the $\Sigma A$ generator incurs practically no overhead.


\section{Evaluation} \label{sec:evaluation}

\subsection{Experimental Methodology} \label{sec:methodology}

\noindent
\textbf{DNN Benchmarks } 
We evaluate seven representative DNN models, including CNNs and transformer networks as summarized in Table \ref{tab:dnn_benchmark}. For CNNs, we evaluate VGG-16, ResNet-34 and ResNet-50 on the ImageNet-1K dataset. For transformers, we choose two vision transformers, ViT-Small and ViT-Base, as well as BERT on MRPC and SST2 tasks from the GLUE dataset \cite{glue}. We obtain pre-trained CNNs and transformers from PyTorch Library and HuggingFace, respectively. 
We then conduct post-training per-channel quantization to obtain the baseline 8-bit models, which shows negligible accuracy loss compared to FP32 models. The 8-bit models are used to evaluate the proposed binary pruning technique and \textit{BitVert} accelerator. 
For every model, we apply two levels of binary pruning, \textit{conservative} (cons) and \textit{moderate} (mod), with a weight group size of 32. For conservative pruning, $10\%$ sensitive channels are maintained at 8 bits and the remaining channels have 2 bit-columns pruned using the rounded averaging strategy. For moderate pruning, $20\%$ sensitive channels are maintained at 8 bits and the remaining channels have 4 bit-columns pruned using the zero-point shifting strategy.
    
\vspace{5pt}
\noindent
\textbf{Accelerator Baselines }
We compare \textit{BitVert} against six DNN accelerators, including four bit-serial accelerators: Stripes~\cite{stripes}, Pragmatic \cite{pragmatic}, Bitlet \cite{bitlet}, BitWave \cite{bitwave}, and two value-based accelerators: SparTen \cite{SparTen}, ANT \cite{ant}. Stripes is an early bit-serial accelerator that exploits reduced precision for DNN computation, yet it mainly relies on 16-bit models and does not consider below-8-bit compression. 
Therefore, we treat Stripes as a dense bit-serial accelerator and use our baseline 8-bit models to evaluate its performance.
Pragmatic and Bitlet target zero-bit skipping during on-chip computation only, while BitWave enhances structured bit-column sparsity to save both computation and memory access. SparTen exploits two-sided value sparsity for DNN acceleration. 
ANT combines different datatypes in a unified manner for low-bit DNN acceleration. We use 6-bit precision to evaluate ANT, a configuration demonstrated by ANT to maintain acceptable accuracy without the need of retraining. 
    
    \begin{table} [b]
      \centering
      \vspace{-5pt}
      \setlength{\tabcolsep}{4pt}
      \renewcommand{\arraystretch}{1.25}
      \footnotesize
        \begin{tabular}{ c | c c c c c c c}
            \Xhline{0.3ex}
              Type & \multicolumn{3}{c|}{CNN} & \multicolumn{4}{c}{Transformer}  \\
            \Xhline{0.3ex}
              Model & 
              \multicolumn{1}{c|}{VGG-16} & 
              \multicolumn{2}{c|}{ResNet-34$\,$/$\,$50} & 
              \multicolumn{2}{c|}{ViT-S$\,$/$\,$B} & 
              \multicolumn{2}{c}{BERT} \\
            \hline
              Dataset & 
              \multicolumn{5}{c|}{ImageNet} & 
              \multicolumn{1}{c|}{MRPC} & 
              \multicolumn{1}{c}{SST2} \\ 
            \hline 
              {FP32 Acc \%} & 
              \multicolumn{1}{c|}{\multirow{1}{*}{73.36}} & 
              \multicolumn{2}{c|}{\multirow{1}{*}{73.31 / 76.13}} & 
              \multicolumn{2}{c|}{\multirow{1}{*}{80.16 / 84.54}} &
              \multicolumn{1}{c|}{\multirow{1}{*}{90.7}}  & \multirow{1}{*}{91.8} \\
            \hline
              {INT8 Acc \%} & 
              \multicolumn{1}{c|}{\multirow{1}{*}{73.35}} & 
              \multicolumn{2}{c|}{\multirow{1}{*}{73.39 / 76.17}} & 
              \multicolumn{2}{c|}{\multirow{1}{*}{80.05 / 84.52}} &
              \multicolumn{1}{c|}{\multirow{1}{*}{90.4}}  & \multirow{1}{*}{91.63} \\
            \Xhline{0.3ex}
        \end{tabular}
        \caption{Summary of evaluated models and datasets.}
      \label{tab:dnn_benchmark}
    \end{table}

\vspace{5pt}
\noindent
\textbf{Implementation } 
    \begin{figure}
        \centering
        \includegraphics[width=1\linewidth]{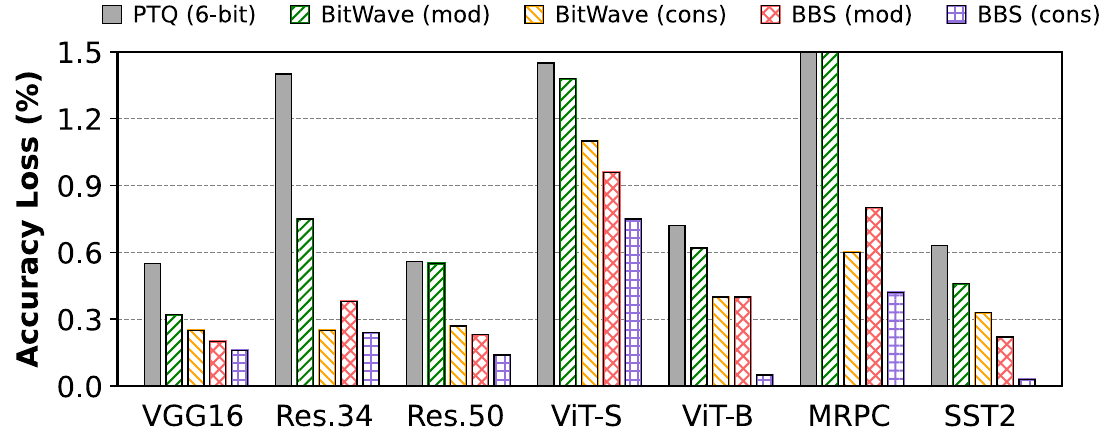}
        \vspace{-12pt}
        \caption{Comparison of accuracy loss between PTQ, BitWave and BBS under conservative (cons) and moderate (mod) compression. }
        \label{fig:accuracy}
        \vspace{-10pt}
    \end{figure}
We implement the proposed binary pruning algotirhm in Pytorch. We design the \textit{BitVert} accelerator at RTL-level using SystemVerilog and synthesize it with Synopsys Design Compiler in TSMC 28nm technology to find area. 
We use Synopsys VCS to generate data-driven activity factors at 800 MHz for power estimation. The area and power of on-chip SRAM buffer are modelled with CACTI \cite{cacti}. To estimate the DRAM power, we use the DDR3 model from DRAMSim3 \cite{dramsim}. For the end-to-end performance evaluation of \textit{BitVert} and other baseline accelerators, we develop cycle-accurate simulators to model the execution time. To ensure a fair comparison, all accelerators are scaled to contain the same number of multipliers, where an 8-bit multiplier is equivalent to eight bit-serial multipliers. For on-chip SRAM, we equip ANT and all bit-serial accelerators with 256 KB activation buffer and 256 KB weight buffer. For SparTen, we reduce the size of its on-chip buffer due to the existence of the local buffer inside every PE.

    \begin{figure*}[t]
        \centering
        \includegraphics[width=1\linewidth]{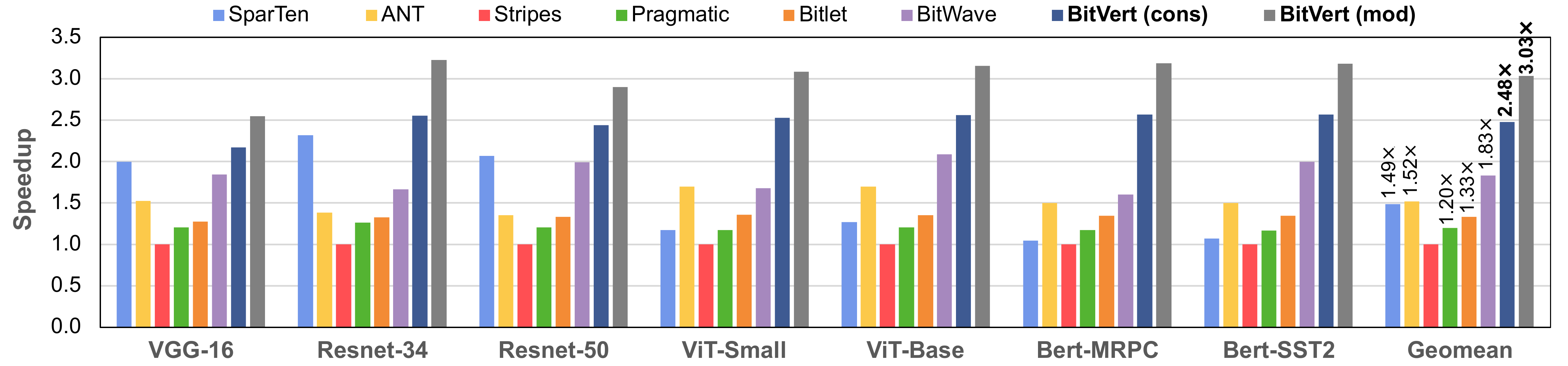}
        \vspace{-15pt}
        \caption{Speedup results normalized to Stripes (higher is better).}
        \label{fig:speedup}
        \vspace{-5pt}
    \end{figure*}

    \begin{figure*}[t]
        \centering
        \includegraphics[width=1\linewidth]{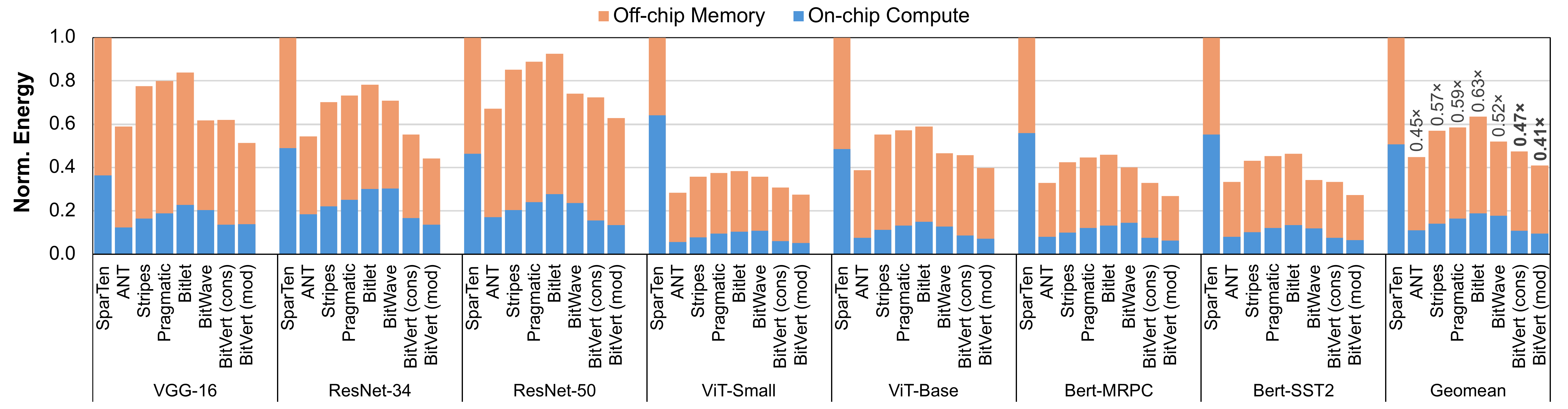}
        \vspace{-12pt}
        \caption{Energy consumption breakdown normalized to SparTen (lower is better).}
        \label{fig:energy}
        \vspace{-10pt}
    \end{figure*}

\subsection{Accuracy Comparison} \label{sec:eval_accuracy}
We first evaluate the accuracy impact of BBS binary pruning compared to naive PTQ and BitWave's bit-flip strategy \cite{bitwave} for compression below 8-bit. 
When using PTQ for compression, we follow the widely-used calibration~\cite{esser2019learned} by calibrating the quantization parameters based on a subset~(1024 images) of the ImageNet dataset. 
In particular, conventional PTQ relies on the calibration dataset to ensure the optimized quantization parameters and accuracy, while the naive data-free quantization leads to significant accuracy degradation~($> 10\%$). 
On the contrary, the proposed BBS compresses the model to lower precision \textbf{without} any calibration dataset.
For both PTQ and BitWave, we use the same setting as BBS by maintaining $20\%$ and $10\%$ sensitive channels for moderate and conservative pruning, respectively. This ensures that our accuracy benefits purely come from the proposed binary pruning. 
    
Fig. \ref{fig:accuracy} shows the accuracy impact of applying different approaches on the baseline DNNs. On average, the conservative and moderate binary pruning can compress the memory footprint of the baseline 8-bit DNNs by $1.29\times$ and $1.66\times$, while incurring an accuracy loss of only $0.25\%$ and $0.45\%$, respectively. 
Both BitWave and BBS with moderate pruning can attain higher accuracy than PTQ. These accuracy improvements stem from their ability to exploit fine-grained bit-level redundancy, thereby preserving more information from the original 8-bit models. Additionally, the proposed binary pruning consistently outperforms BitWave. This is because BBS allows any bit significance to be zero or one, thus retaining all quantization levels of the 8-bit precision. 

\vspace{5pt}
\noindent
\textbf{Comparison against ANT } 
    \begin{table} [b]
      \centering
      \vspace{-5pt}
      \setlength{\tabcolsep}{8pt}
      \renewcommand{\arraystretch}{1.25}
      \footnotesize
        \begin{tabular}{ c | c | c  }
            \Xhline{0.3ex}
              Model & \multicolumn{1}{c|}{\textbf{BBS (mod)}} & \multicolumn{1}{c}{ANT \cite{ant}} \\
            \Xhline{0.3ex}
              VGG-16 & \multicolumn{1}{c|}{\textbf{0.2\% (4.32 bits)}} & 0.68\% (6 bits)
             \\
            \hline
              ResNet-50 & \multicolumn{1}{c|}{\textbf{0.23\% (4.79 bits)}} & 0.89\% (6 bits)
             \\
            \Xhline{0.3ex}
        \end{tabular}
        \caption{Comparison of accuracy loss and weight bit width between BBS and 6-bit ANT without fine-tuning. }
      \label{tab:ant_comparison}
    \end{table}
We compare the accuracy between moderate binary pruning and ANT \cite{ant}. As shown in Table~\ref{tab:ant_comparison}, BBS outperforms ANT in terms of both accuracy and effective weight bit width. While ANT uses adaptive datatypes for low-bit quantization, it cannot take the advantage of inherent bit-level redundancy. On the other hand, the binary pruning fully exploits the bit-level sparsity to best preserve the original 8-bit weight distribution, resulting in minimal accuracy degradation.

\vspace{5pt}
\noindent
\textbf{Comparison against PTQ Works } 
We compare the accuracy loss between BBS and state-of-the-art PTQ works, including Microscaling \cite{microscaling} and NoisyQuant \cite{noisyquant}, on vision transformers. We apply 6-bit weight quantization using the two PTQ methods while maintaining activation to 8-bit. Table~\ref{tab:ptq_comparison} shows that the moderate binary pruning outperforms NoisyQuant with lower memory footprint. Moreover, the conservative binary pruning has much better accuracy than Microscaling at similar bit width. Miscroscaling also has an \text{8-bit} metadata, which represents the shared exponent for a group of 32 weights. However, the exponent is determined by the largest value in every group, which forces small values to become zero due to insufficient operand precision to store the aligned mantissa. On the other hand, BBS exploits bit-level redundancy to better preserve the statistical characteristics of uncompressed weight, thereby achieving higher accuracy.

    \begin{table} [b]
      \centering
      \vspace{-5pt}
      \setlength{\tabcolsep}{8pt}
      \renewcommand{\arraystretch}{1.25}
      \footnotesize
        \begin{tabular}{ c | c | c | c | c}
            \Xhline{0.3ex}
              & \multicolumn{2}{c|}{ViT-Small} & \multicolumn{2}{c}{ViT-Base} \\
            \cline{2-5}
              & $\Delta\,$Acc $\downarrow$ & Bits & $\Delta\,$Acc $\downarrow$ & Bits \\
            \Xhline{0.3ex}
              Microscaling \cite{microscaling} & $2.49\%$ & 6.25 & $0.33\%$ & 6.25 \\
            \hline
              NoisyQuant \cite{noisyquant} & $2.08\%$ & 6 & $0.64\%$ & 6 \\
            \hline
              BBS (cons) & $0.75\%$ & 6.33 & $0.05\%$ & 6.25 \\
            \hline
              BBS (mod) & $0.96\%$ & 5.19 & $0.39\%$ & 5.07 \\
            \Xhline{0.3ex}
        \end{tabular}
        \caption{Comparison of accuracy loss and weight bit width between BBS, Microscaling and NoisyQuant. }
      \label{tab:ptq_comparison}
    \end{table}

\subsection{Accelerator Performance and Energy} 
\noindent
\textbf{Performance } 
Fig. \ref{fig:speedup} presents the accelerator performance normalized to that of Stripes. On average, \textit{BitVert} with conservative and moderate binary pruning achieves $2.48\times$ and $3.03\times$ speedup compared to Stripes, respectively. 
These speedups are attributed to exploiting both balanced BBS and binary pruning for abundant bit skipping and reduced memory access. Despite leveraging two-sided value sparsity, SparTen demonstrates limited performance on transformer-based models due to the lack of weight value sparsity in 8-bit models and nearly-dense activations from non-ReLU functions. ANT only explores reduced value precision but not fine-grained bit-level sparsity, leading to $1.63\times$ and $1.97\times$ lower speedup than \textit{BitVert} at conservative and moderate pruning, respectively. 
While Pragmatic and Bitlet utilize variable degrees of bit-level sparsity, they suffer from workload imbalance and lack of exploration in further compressing DNNs below 8-bit. 
This explains why \textit{BitVert} outperforms Pragmatic and Bitlet by $1.86$ -- $2.53\times$ across all benchmarks. Although BitWave exploits structured bit-column pruning to achieve better performance, its moderate pruning results in unacceptable accuracy loss ($> 1\%$) on many DNNs such as ViT-small and Bert-MRPC. Therefore, it has to reduce the degree of pruning for improved accuracy while sacrificing performance. Overall, \textit{BitVert} provides the best accuracy-performance trade-offs, with up to $1.98\times$ speedup over BitWave.

\vspace{5pt}
\noindent
\textbf{Energy Consumption } 
Fig. \ref{fig:energy} presents the normalized energy breakdown of different accelerators. where the on-chip compute energy includes both buffer and core energy. 
SparTen demonstrates the poorest energy efficiency primarily due to its substantial overhead from the sparse bitmask encoding ($12.5\%$ at 8-bit precision) and the expensive hardware required to exploit sparsity. This overhead is particularly pronounced in 8-bit DNNs, where value sparsity is inherently scarce. As a result, SparTen consumes $2.13\times$ and $2.44\times$ higher energy than \textit{BitVert} with conservative and moderate pruning, respectively. Although ANT is able to quantize both activations and weights, it dissipates higher energy than \textit{BitVert} with moderate pruning due to the complicated hardware to support custom data types.
Owing to the balanced BBS-skipping and substantial reduction in model size, \textit{BitVert} with moderate pruning achieves an average energy reduction of $1.39\times$, $1.43\times$, $1.54\times$, and $1.27\times$ over Stripes, Pragmatic, Bitlet, and BitWave, respectively. 

\subsection{Analysis of Load Imbalance}
    \begin{figure}
        \centering
        \includegraphics[width=1\linewidth]{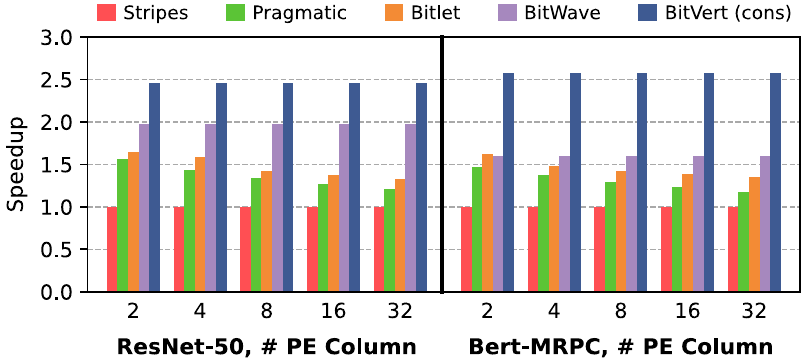}
        \vspace{-10pt}
        \caption{Normalized speedup on ResNet-50 and Bert-MRPC with increasing number of PE columns (\textit{i.e.}, processing more weight groups in parallel).}
        \label{fig:scability}
        \vspace{-5pt}
    \end{figure}
    
\textit{BitVert} can leverage the structured BBS for improved load balance. 
Fig. \ref{fig:scability} demonstrates this with the performance on ResNet-50 and Bert-MRPC with respect to different number of PE columns, where every PE column processes a different weight group.
When there are more PE columns, Pragmatic and Bitlet exhibit a noticeable drop in speedup over Stripes that does not exploit bit sparsity. For instance, when the number of PE columns increases from 2 to 32, the speedup of Bitlet on Bert-MRPC drops from $1.63\times$ to $1.35\times$. This is because that processing more weight groups in parallel exacerbates the load imbalance across PE columns, and the performance is bottlenecked by the weight group with the lowest bit sparsity. In contrast, the structured bit sparsity allow BitWave and \textit{BitVert} to efficiently scale the performance, thus maintaining nearly constant speedup over Stripes. Moreover, \textit{BitVert} always achieves the highest performance thanks to the binary pruning that can induce higher BBS with negligible accuracy loss.

\begin{figure}
        \centering
        \includegraphics[width=1\linewidth]{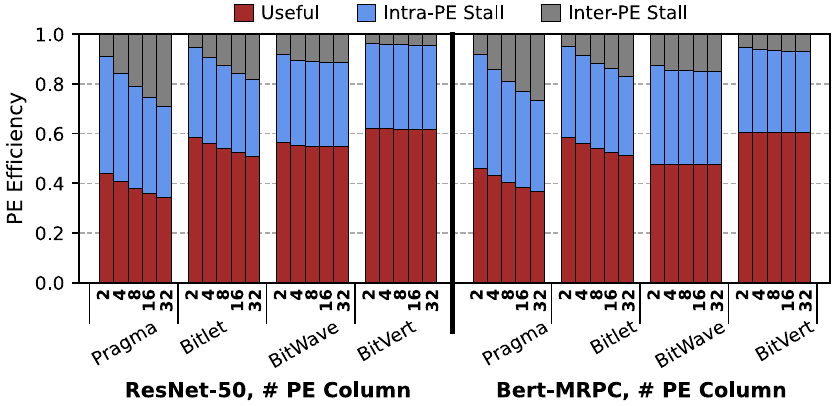}
        \vspace{-12pt}
        \caption{Breakdown of execution cycles w.r.t. the number of PE columns.}
        \label{fig:load_imbalance}
        \vspace{-5pt}
    \end{figure}

Fig. \ref{fig:load_imbalance} further details the breakdown of execution time with respect to the number of PE columns to highlight its impact on load balance. Since one PE contains many bit-serial multipliers, intra-PE stall can be caused by a multiplier that needs to process more effectual bits. On the other hand, the inter-PE stall arises from variance in bit sparsity across different weight groups. As the number of PE columns increases, Pragmatic and Bitlet experience higher intra-PE and inter-PE loss, which explains their lower resulting speedup. BitWave only exploits coarse-grained bit-column sparsity that has much lower occurrence than fine-grained BBS. Therefore, it shows lower PE utilization than \textit{BitVert}. Furthermore, \textit{BitVert} has minimal inter-PE stall due to the more balanced distribution of BBS across different weight groups, thereby achieving superior performance over other bit-serial accelerators. 

    \begin{table} [b]
      \centering
      \setlength{\tabcolsep}{4pt}
      \renewcommand{\arraystretch}{1.25}
      \footnotesize
        \begin{tabular}{ c | c c | c c }
            \Xhline{0.3ex}
              \multirow{1}{*}{Sub-group} & \multicolumn{2}{c|}{Without Optimization} & \multicolumn{2}{c}{With Optimization} \\
            \cline{2-5}
              {Size} & Area ($um^2$) & Power ($mW$) & Area ($um^2$) & Power ($mW$) \\
            \Xhline{0.3ex}
              16 & 1342.3 & 0.61 & 971.5 & 0.53 \\
            \hline
              8 & 896.6 & 0.49 & 739.6 & 0.45 \\
            \hline
              4 & 878.7 & 0.51 & 786.5 & 0.47 \\
            \Xhline{0.3ex}
        \end{tabular}
        \caption{PE area and power of \textit{BitVert} with different sub-group sizes before and after applying our circuit optimizations.}
      \label{tab:pe_sub_group}
    \end{table}

\subsection{PE Design Space Exploration} \label{sec:hardware_explore}
Recall from Section \ref{sec:bitvert_pe} that the sub-group size within the \textit{BitVert} PE offers a trade-off between area and power. A smaller sub-group has lower mux cost, but increases the number of subtractors. Furthermore, by exploiting the structured nature of BBS and its encoding scheme, we are able to further reduce the PE area by using compact mux and a smaller BBS multiplier. Hence, we conduct a PE design space exploration to evaluate the optimal group size and the proposed optimizations. As shown in Table \ref{tab:pe_sub_group}, a sub-group size of 16 without optimization incurs a significant area overhead of $38.2\%$ compared to the optimized design. In the end, a sub-group size of 8 with the proposed PE optimization offers the best trade-off between area and power, which is therefore adopted in our \textit{BitVert} accelerator.

\subsection{PE Area and Power Comparison}
The \textit{BitVert} accelerator adopts an area- and energy-efficient PE with low overhead to support BBS. We compare the PE design of \textit{BitVert} and other bit-serial accelerators, with all PEs containing 8 bit-serial multipliers at 800 MHz target frequency. Table \ref{tab:pe_comparison} summarizes the area and power of different PEs. 
Bitlet experiences the highest area and power consumption due to significant overhead (\textit{e.g.}, a 64-1 mux before every bit-serial multiplier) for zero bit skipping. 
Pragmatic needs a variable shifter to align the bit significance, leading to a larger bit-serial multiplier and non-trivial overhead. BitWave requires 2's complementer to support sign-magnitude arithmetic, resulting in $1.32\times$ larger area and $1.4\times$ power than Stripes. Moreover, since BitWave can only leverage coarse-grained bit-column sparsity, the potential performance improvement is limited. 
The proposed \textit{BitVert} enjoys the optimal trade-off between performance and hardware cost. Its PE occupies $1.39\times$ area and consumes $1.22\times$ power compared to Stripes, yet is able to exploit $50\%$ balanced BBS and binary pruning for efficient bit skipping and model compression, respectively. 
Since BBS naturally exists in a bit-vector with arbitrary length and does not depend on the operand precision, it provides a promising solution for future bit-serial computing paradigm. 

    \begin{table} [b]
      \centering
      \setlength{\tabcolsep}{5pt}
      \renewcommand{\arraystretch}{1.25}
      \footnotesize
        \begin{tabular}{ c | c c c c | c }
            \Xhline{0.3ex}
              \multirow{2}{*}{Accelerator} & \multicolumn{4}{c|}{PE Area ($um^2$)} & \multicolumn{1}{c}{\multirow{1}{*}{PE Power}} \\
            \cline{2-5}
              & \multicolumn{1}{c}{Multiplier} & \multicolumn{1}{c}{Others} & \multicolumn{1}{c}{Total}  & \multicolumn{1}{c|}{Ratio} &  ($mW$) \\
            \Xhline{0.3ex}
              Stripes \cite{stripes} & 286.3 & 246.5 & 532.8 &  $1\times$ & 0.37 \\
            \hline
              Pragmatic \cite{pragmatic} & 319.2 & 603.9 & 923.1 & $1.73\times$  & 0.51  \\
            \hline
              Bitlet \cite{bitlet} & 223.2 & 1442.4 & 1665.6 & $3.13\times$ & 0.57 \\
            \hline
              BitWave \cite{bitwave} & 286.3 & 416.1 & 702.4 & $1.32\times$ & 0.49 \\
            \hline
              \textbf{BitVert (ours)} & \textbf{332.4} & \textbf{407.2} & \textbf{739.6} & \textbf{1.39$\times$} & \textbf{0.45} \\
            \Xhline{0.3ex}
        \end{tabular}
        \caption{PE area and power of BitVert and prior bit-serial accelerators under 28 nm technology and 800 MHz frequency. }
      \label{tab:pe_comparison}
    \end{table}

\subsection{Accuracy-Efficiency Trade-offs}
    \begin{figure}[t]
        \centering
        \includegraphics[width=1\linewidth]{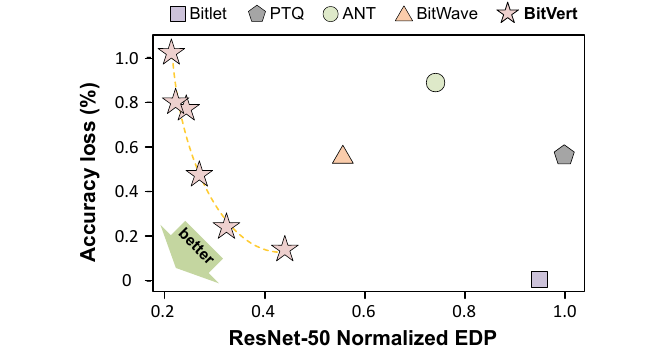}
        \vspace{-12pt}
        \caption{EDP-acccuracy loss pareto frontier for ResNet50.}
        \label{fig:pareto}
        \vspace{-10pt}
    \end{figure}
The proposed binary pruning and \textit{BitVert} can offer good trade-offs between accuracy and efficiency. To demonstrate this, we conduct design-space exploration on ResNet-50 with different pruning ratios. We compare the relationship between energy-delay product (EDP) and accuracy loss of \textit{BitVert} and previous works, including Bitlet, BitWave, ANT and conventional PTQ. As shown in Fig. \ref{fig:pareto}, the lower left region indicates a good trade-off between accuracy and EDP. Although BitWave and ANT propose different algorithm-hardware co-design approaches for DNN compression and acceleration, they fail to preserve the original value distribution of the baseline model and do not efficiently leverage the balanced bit sparsity that inherently appears in DNNs. In contrast, binary pruning is able to preserve all quantization levels of the original DNN. Combining with BBS and efficient hardware design, \textit{BitVert} is able to always sit on the Pareto frontier.

\subsection{Applicability to Large Language Models}
\begin{figure}[t]
        \centering
        \includegraphics[width=1\linewidth]{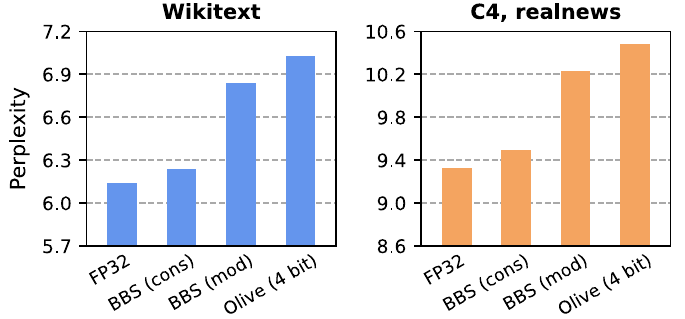}
        \vspace{-12pt}
        \caption{Comparison between BBS and Olive on compressing Llama-3-8B weights. The accuracy metric is perplexity, \textbf{lower is better}.}
        \label{fig:llm_ppl}
        \vspace{-10pt}
    \end{figure}

Large language models (LLMs) have achieved great success in generative tasks \cite{llama, opt}. 
We compare BBS with a recent PTQ work Olive \cite{olive} for LLM weight compression. We evaluate a state-of-the-art LLM, Llama-3-8B \cite{llama-3} on Wikitext~\cite{wikitext} and C4 \cite{c4-dataset} datasets. For BBS, we apply conservative and moderate binary pruning to \textit{all} weight channels with a group size of 32, resulting in an effective weight precision of 6.25 and 4.25 bits, respectively. Fig. \ref{fig:llm_ppl} shows the accuracy impact of different compression methods. The moderate BBS pruning achieves better perplexity than Olive with a similar memory footprint (4.25 vs. 4 bits), while the conservative BBS pruning has little perplexity loss compared to the FP32 baseline. To compare the hardware efficiency, we synthesize the Olive PE for 4-bit weight and 8-bit activation. Table \ref{tab:pe_comparison_olive} shows that the proposed \textit{BitVert} PE with moderate binary pruning can achieve $1.58\times$ better performance per area compared to Olive. The benefits of \textit{BitVert} are twofold. First, Olive adopts separate datatypes for normal and outlier values, where the latter has a much wider numerical range. Therefore, the Olive PE requires a larger multiplier than fixed-point PE to accommodate outliers. Second, the \textit{BitVert} PE exploits BBS to efficiently compute 16 multiplications in 4 cycles under moderate pruning, while the Olive PE does not leverage bit sparsity and only computes one multiplication per cycle.

    \begin{table} [t]
      \centering
      \vspace{10pt}
      \setlength{\tabcolsep}{8pt}
      \renewcommand{\arraystretch}{1.25}
      \footnotesize
        \begin{tabular}{ c | c | c | c | c }
            \Xhline{0.3ex}
              \multirow{2}{*}{Accelerator} & Area & Power & Norm. & Norm. \\
              & ($um^2$) & ($mW$) & Perf & Perf / Area \\
            \Xhline{0.3ex}
              Olive \cite{olive} & 291.6 & 0.18 & 1$\times$ & 1$\times$ \\
            \hline
              BitVert (mod) & 739.6 & 0.45 & 4$\times$ & 1.58$\times$ \\
            \Xhline{0.3ex}
        \end{tabular}
        \caption{Comparison between Olive and \textit{BitVert} PEs.}
        \vspace{-10pt}
      \label{tab:pe_comparison_olive}
    \end{table}
    

\section{Conclusion}
In this paper, we introduce BBS, a new concept to exploit bit-level sparsity in a symmetrical way to prune either zero-bits or one-bits. BBS pushes the limit of post-training DNN compression to a new state-of-the-art through binary pruning, a data-free optimization that generates bi-directional sparse bit columns inside DNN weights while maximally preserving the statistical characteristics of the original uncompressed model. As a result, the proposed binary pruning technique achieves much higher accuracy compared to previous bit-sparsity-aware pruning methods. On top of the algorithmic innovation, we design a bit-serial accelerators named \textit{BitVert} with an area- and power-efficient PE to fully mine the potential of BBS. Compared to prior DNN accelerators, \textit{BitVert} achieves up to $3.03\times$ speedup and $2.44\times$ energy saving, while having negligible accuracy degradation on both vision and language models with large-scale benchmark datasets.

\section*{Acknowledgment}
This project is supported in part by Intel Corporation 
and the Center for the Co-Design of Cognitive Systems (CoCoSys) in JUMP 2.0, an SRC Program sponsored by DARPA.
We would like to thank Mahesh Iyer, Grace Zgheib, Sergey Gribok, Ahmed AbouElhamayed, Zhewen Yu, Marta Andronic, and the anonymous reviewers for their constructive feedback. We also thank Man Shi for the helpful discussion about BitWave. The code for BBS binary pruning can be found at \texttt{https://github.com/yc2367/BBS-MICRO.git}.

\bibliographystyle{IEEEtranS}
\bibliography{refs}

\end{document}